\documentclass[12pt]{article}

\usepackage{arxiv}
\usepackage{graphicx,url}
\usepackage[utf8]{inputenc}
\usepackage{mathrsfs}   

\title{A Preliminary Study for Literary Rhyme Generation based on Neuronal Representation, Semantics and Shallow Parsing
}

\author{Luis-Gil Moreno-Jiménez\\
  Laboratoire Informatique d'Avignon \\
  France\\
   \texttt{\small luis-gil.moreno-jimenez@univ-avignon.fr} \\
 
   \And
  Juan-Manuel Torres-Moreno\\
  Laboratoire Informatique d'Avignon \\
  France\\
  \texttt{\small juan-manuel.torres@univ-avignon.fr} \\  
  
  \AND
  Roseli~S.~Wedemann \\
  Inst. de Matem\'atica e Estatística -- Universidade do Estado do Rio de Janeiro \\
    RJ Brazil \\
  \texttt{\small roseli@ime.uerj.br} \\  
}

\date{}

\begin{document} 

\maketitle

\begin{abstract}
In recent years, researchers in the area of Computational Creativity have studied the human creative process 
proposing different approaches to reproduce it with a formal procedure. 
In this paper, we introduce a model for the generation of literary rhymes in Spanish, combining 
structures of language and neural network models 
The results obtained with a manual evaluation of the texts  generated by our algorithm are encouraging.
\end{abstract}


\section{Introduction}
For many years, research in Artificial Intelligence (AI) has directed efforts towards automating processes to perform specific academic, industrial or economic tasks for society. However, the investigation and development of procedures for the automation of human artistic and creative processes has not had as much attention due to the complexities involved in these activities. Procedures developed for these purposes involve mathematical-computational methods designed to process and learn from a large quantity of digital data, so as to detect patterns in order to simulate the creative process (CP), as explained by Boden in \cite{boden2004creative}. 

In this paper, we introduce a model for the generation of rhymes with literary components. Our proposal is based on findings detailed in \cite{moreno_nldb}, where Automatic Text Generation (ATG) techniques are combined with neural network (NN) based models, such as the {\it Word2vec\/} algorithm \cite{mikolov2013linguistic}, for the generation of literary texts. In Section \ref{sec:rel_work}, we present some of the literature regarding literary text generation, focusing on methods related to this paper. In Section \ref{sec:rimax}, we explain the RIMAX model used to generate the rhyming words.
In Section \ref{sec:corpora}, we describe the corpora used for the learning phase of our models. 
Then, in Section \ref{sec:model}, we explain the methodology implemented for the generation of rhymes. 
We show some experiments and examples in Section \ref{sec:exp}, as well as the results of evaluations conducted by humans. Finally, we present conclusions and propose possible future works in Section \ref{sec:conclusion}.

\section{Related Work}
\label{sec:rel_work}

There has been much interest and work in the area of ATG with different and interesting goals, regarding the different types of texts. Many of the proposed algorithms are based on neural networks.
In \cite{checklist_ATG}, the authors generate coherent text using a recurrent neural network (RNN) and a {\it neural checklist model}. 
Their RNN predicts the best context from a list of keywords. 
Another RNN approach is proposed in \cite{clark-etal-2018-neural} for generating narrative text, such as fiction or news stories. Entities mentioned in the
text are represented by vectors, which are updated as the text generation proceeds as they represent different contexts and guide the RNN in determining the vocabulary to be retrieved in order to generate a narrative.

We note that there are other ATG techniques, such as {\it text realization} that creates text in a human language, {\it e.g.} English or French, from a syntactic representation \cite{molins2015jsrealb}. 
%
Oliveira has written a survey of work treating the automatic generation of poetry \cite{oliveira2017survey}, and presents his own method for generating poems based on the use of templates (\textit{canned text}) in \cite{Oliveira2015}.
Another work based on canned text is presented in \cite{agirrezabal2013pos}, which generates strophes of verses for Basque poetry. 
In \cite{Zhang2014Poetry}, a RNN was proposed for the generation of Chinese poetry based on learning of known text structure. 

\section{Semantic Rhyme}
\label{sec:rimax}

RIMAX is the first automatic system for detecting semantic rhymes in Spanish \cite{torres_rimax}. It contains the following ingredients: 1) a rhyming dictionary, 2) the set of definitions of those rhymes and 3) a strategy to measure semantic proximity. 
This procedure can be applied to different romance languages, although we have chosen the Spanish language spoken in Mexico, given the availability of some useful resources and tools, such as the Dictionary of Mexican Spanish (DEM)\footnote{\textit{Diccionario del español de M\'exico}, \url{https://dem.colmex.mx/}.} and the Rhyming Dictionary  \cite{rem:18}.



Rhyming dictionaries gather words according to rhyming patterns. 
 \textit{Consonant} rhymes share ending sequences of vocalic and consonant sounds and \textit{assonant} rhymes share similar vowel sounds. These two classes are thus based on pronunciation features, not on writing patterns. 
Also, since consonance and assonance depend on the stressed syllable, words which end with a stressed syllable are grouped together, those whose stressed syllable is the next to last appear together, and so on\footnote{For example, the penultimate syllables of the following Spanish words are the stressed syllables: \textit{angula, chula, mula, chamula}. So these words should appear together in a rhyming dictionary.}.
%
%
In this paper we have used the nomenclature of the DEM to automatically generate a phonological transcription. 

\subsection{Rhyme Ranking by Definition Similarity}
\label{sec:method}

Online dictionaries 
offer useful and simple advantages such as ranking and ordering of results. From a language perspective, it is interesting that text mining techniques can be applied to accomplish this. 
In fact, text similarity measures can be used to determine how similar word definitions are, i. e. measuring definition similarity.  
 
Let $D$ be a dictionary containing the set of defined words $w$ and the set of definitions $d$. 
Since a word may have several senses, let $d_{ij}$ be the $j$th definition of word $w_i$ in $D$. 
Similarly, let $d'_{kl}$ be the $l$th definition of word $w_k$. 
Also, let $\vec{v_{ij}}$ and $\vec{v'_{kl}}$ be vectors where the frequencies of lemmatized or ultrastemmized \cite{DBLP:journals/corr/abs-1209-3126} content words of definitions $d_{ij}$ and $d'_{kl}$ are stored.
Then, the similarity between $d_{ij}$ and $d'_{kl}$ can then be measured using the well-known quantity called {\it cosine similarity measure}, $s_c$.

In order to find semantic rhymes, each member of the rhyming set of word $x$ will be weighed according to how similar its definition is to that of $x$, using the similarity measurement $s_c$. 
Hence, given a query word, consonance and assonance lists are generated and ordered by the calculated similarity among definitions. 
The program RIMAX allows us to select some parameter values for experiments.


\section{Corpus}
\label{sec:corpora}

The corpus \textbf{MegaLite-Es} was used to train our model. It consists of $5~075$ literary documents (mainly books) in Spanish.
This corpus can be useful for different NLP tasks.
The documents of \textbf{MegaLite-Es} were obtained from different personal collections and, for copyrights reasons, the distribution of the original documents is not possible. Instead of this, in \cite{megalite} the authors propose some alternative resources.


\subsection{Corpus Structure}
\label{sec:structure}

The 5 075 documents in \textbf{MegaLite-Es} were written by 1~336 Spanish-speaking authors and official translations from languages other than Spanish. The documents represent different literary genres such as plays, poems, tales, essays, etc. We thus consider that this corpus is suitable for training {\it Word2vec\/} models.  


The original documents, in heterogeneous formats\footnote{pdf, txt, html, doc, docx, odt, etc.} were processed to be converted into \textit{utf8} encoded documents. A segmentation process divided the texts into sentences, corresponding to regular expressions, using a tool developed in PERL 5.0. Some undesirable data like: mutilated words, strange symbols and an unusual disposition of paragraphs could not be treated, although these situations are usual when dealing with this kind of corpora.
Some characteristics of \textbf{MegaLite-Es} are detailed in Table~\ref{tab:1}.


\begin{table}[htbp]
  \centering
  \caption{Characteristics of \textbf{MegaLite-Es} corpus (M = $10^6$ and K = $10^3$).}
  \begin{tabular}{|l|c|c|c|c|c}
     \hline
   \textbf{MegaLite-Es} & \textbf{Sentences} & \textbf{Tokens} & \textbf{Characters} & \textbf{Authors} \\
    \hline
   {\bf Overall}   & 15 M & 212 M & 1 265 M & 1 328 \\ \hline
   {\bf Avg per document} & 3 K & 41.8 K & 250 K & -- \\
    \hline
  \end{tabular}
  \label{tab:1}
\end{table}

\textbf{MegaLite-Es} has the advantage of being very extensive and suitable for automatic learning.
It has, however, the disadvantage that many of its sentences consist of general language, without literary elements (stylized vocabulary or literary figures). However, these sentences often allow for fluent reading and provide the necessary links between the ideas expressed in the text, although they could imply some noisy results. As numbers identifying pages, chapters, sections or index could imply errors in the detection of sentences during segmentation,
a manual process was performed to remove this undesirable data, although these errors may occur in a linguistic corpus with unstructured text.


\section{Text Generation Model}
\label{sec:model}

In this section, we describe the model we have proposed for the generation of literary rhymes. The model consists of two steps described as follows.

\subsection{First Step: Canned Text Method}
\label{sec:FirstStep}

We implemented a \textit{Canned Text} method, which has the advantage of being efficient for syntactic analysis in ATG tasks \cite{Templatebased}, to generate grammatical templates named Partially Empty Grammatical Structures (PGSs). 
Each PGS is composed of Part-of-Speech (POS) tags\footnote{A POS tag indicates the part of speech grammatical category of a word.} and function words\footnote{Prepositions, pronouns, auxiliary verbs, or conjunctions.}. The POS tags are retrieved by Freeling \cite{freeling}.
PGSs are created from a template set, called {\it TempSet}, that consists of sentences selected manually from the \textbf{MegaLite-Es} corpus, according to the following rules.
\begin{itemize}
    \item Each sentence must express a concrete idea.
    \item Each sentence must have a length $N$, such that $5\leq N \leq 10$. 
    \item The sentence should contain at least three lexical words\footnote{Verbs, adjectives, nouns and adverbs}.
\end{itemize}

For generating rhymes, the process begins by selecting two original sentences $f1$ and $f2$ from {\it TempSet} with length $N$, $5\leq N \leq 10$. Sentences $f1$ and $f2$ must satisfy two additional conditions: (\textit{1}) both sentences must finish with a lexical word; (\textit{2}) the lexical words finishing the sentences must have the same grammatical inflection.
These sentences are analyzed with FreeLing to detect lexical words that are replaced by POS tags. We concentrate on lexical words because they provide the most meaningful information in a text \cite{bracewell2005multilingual}. Function words are retained in the sentence, as these are useful for maintaining the grammatical coherence and we therefore do not change them. 

\begin{figure}[htbp]
\centering
  \includegraphics[width=15cm ]{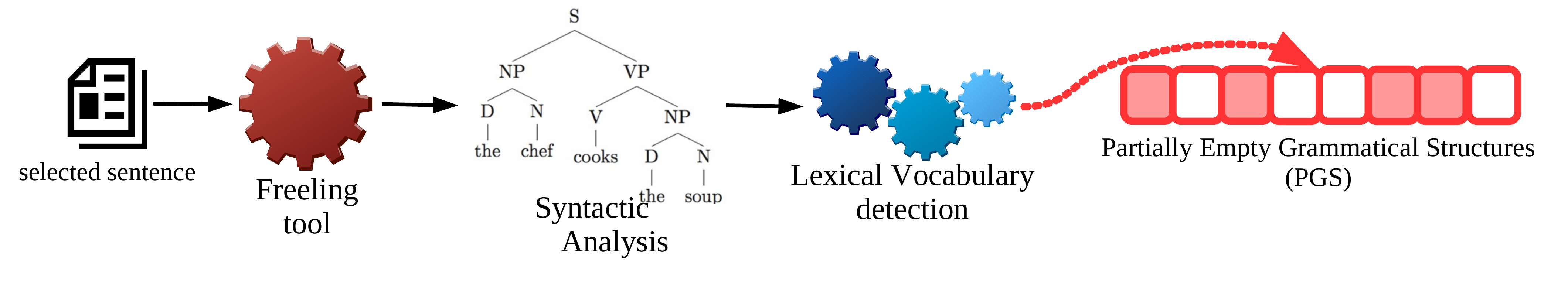}
  \caption{First step: Canned Text Method}
  \label{fig:textinlata}
\end{figure}

The idea is to generate artificial sentences from ``human'' sentences, respecting their grammatical structure and substituting only the lexical words by words with the same linguistic inflection but a different meaning. This technique of text generation is well-known as homo-syntax. In contrast to the paraphrase that keeps the same meaning between the original and the generated texts and changes the grammar, homo-syntax seeks to generate a new text with a different meaning than the original text, although with the same grammatical structure.
In Fig. \ref{fig:textinlata}, we show an illustration of the proposed model. The filled boxes represent function words, whereas the empty boxes represent the lexical words that are replaced by POS tags. Once the pair of sentences has been transformed into a PGS, it will be further changed by the procedure of the second step.


%


\subsection{Second step: Vocabulary selection}

In this step, the POS tags in the PGSs are replaced by a vocabulary selected with the \textit{Word2vec}\footnote{{\it Word2vec\/} belongs to a group of ANN models, used to produce embeddings \cite{bengio2013representation}.} model. In a trained \textit{Word2vec} model, each word, $j$, in the vocabulary is represented by a vector $\vec{L}_j$ with   numerical valued elements. This allows for the implementation of different mathematical procedures such as, for example, interpreting a semantic relation between a pair of words by calculating the cosine similarity between their vectors. These numerical vector representations are called \textit{embeddings}.

The hyper-parameters configured for the \textit{Word2vec} training were: {\it Iterations} = \textbf{10} (the number of training epochs over the {\bf MegaLite-Es} corpus), {\it Minimum count\/} = \textbf{3} (the minimum number of times that a word must appear in the corpus to be included in the model's vocabulary), {\it Vector size\/} = \textbf{100} (the dimension of vectors, the \textit{embeddings}) and {\it Window size\/} = \textbf{5} (the radius of adjacent words that will be related to the current word within a sentence, during the training phase of the model). We trained the model following the skip-gram procedure \cite{mikolov2013efficient}. Using the \textbf{MegaLite-Es} corpus for training, a trained model of 346~616 \textit{Embeddings}\footnote{Term used for the numerical representation of words for NLP, typically in the form of a real-valued vector that encodes the meaning of the word.} is obtained.

\subsubsection{Word2vec Model}
For the replacement, we used the analogical reasoning task introduced in \cite{mikolov2013efficient}. This reasoning consists of considering the relation between words, e. g. ``France'', ``Paris'', ``Spain'' and a missing word $x$. We suppose that ``France'', ``Paris'' and ``Spain'' are words that belong to the vocabulary of a corpus {\bf CorpA} that was used to train {\it Word2vec}, and therefore,  $\vec{Paris}, \vec{France}$, and $\vec{Spain}$ are the corresponding vectors associated to these words after training, respectively. The word $x$ is then determined by finding a vector $\vec{x}$ associated to a word in {\bf CorpA},  such that $\vec{x}$ is closest to $\vec{y}$ = $\vec{Paris} - \vec{France} + \vec{Spain}$, according to the cosine similarity between $\vec{y}$ and $\vec{x}$ (see Eq. (\ref{eq:cosinesim})). This specific example is considered to have been answered correctly, if $\vec{x}$ is the vector corresponding to ``Madrid'' in the vocabulary of {\bf CorpA}. 
For the replacement of POS tags, we consider the three following words, their embedding vectors and Eq. (\ref{eq:reasoning}),
\begin{itemize}
    \item[$Q$:] the context given by the user as a single query word,
    \item[$O$:] the original word in $f1$ or $f2$ that is replaced by the POS tag,
    \item[$A$:] the word adjacent to $O$ on the left, in sentence $f1$ or $f2$, if it exists,
\end{itemize}
\begin{equation}
  \label{eq:reasoning}
  \vec{y} =  \vec{A} - \vec{O}  + \vec{Q} \, ,
\end{equation}
where $\vec{y}$ is the vector which we will use to choose the $M=4~000$ closest embeddings. We rank the $M$ embeddings in a list $\mathscr{L}$, by calculating the cosine similarity between the $j^{th}$ embedding, $\vec{L_j}$, and $\vec{y}$,
\begin{equation}
    \label{eq:cosinesim}
    \theta_j = \cos(\vec{L_j},\vec{y}) = \frac{\vec{L_j} \cdot \vec{y}}{||\vec L_j|| \cdot ||\vec y||} \,\,\,\, 1 \le j \le M .
\end{equation}
$\mathscr{L}$ is ranked according to decreasing $\theta_j$.
%
%
%

If we are replacing the first POS tag, then $A=None$, so we only compute $\vec{y} = \vec{O} + \vec{Q}$. For example, for the {\it Query} word {\it love} and the sentence: \textit{I play the guitar}, we will replace the verb \textit{play} and the noun \textit{guitar}. Starting by the verb \textit{play}, we compute $\vec{y} = \vec{play} + \vec{love}$ to get the ranked list $\mathscr{L}$. Some examples of returned embeddings are: \textit{to like, to role, enchanting} and {\it abandon}. This list is then used in the analysis performed by the language model based on bigrams.


\subsubsection{Language Model Analysis (Bigrams)}
This step consists of calculating the conditional probability of a word, given a preceding word, that is 
\begin{equation}
    P(w_n|w_{n-1})=\frac{P(w_n \land w_{n-1})}{P(w_{n-1})}  \, .
\end{equation}
Each bigram in the \textbf{MegaLite-Es} corpus has been detected and computed in this way. As a result, we have a new list of bigrams, $LB$. The bigrams are composed only by lexical and function words, ignoring punctuation, numbers and symbols.
For each element in $\mathscr{L}$, we configure two bigrams, as $b_1$ and $b_2$, where:
\begin{itemize}
    \item $b_1$ is the adjacent word to the left of $O$ in $f1$ or $f2$, concatenated with the current analyzed word, $L_j$, and
    \item $b_2$ is the current analyzed word $L_j$ concatenated with the adjacent word to the right of $O$, in  $f1$ or $f2$.
\end{itemize}
We calculate the arithmetic mean, $bm$, of the frequencies of occurrence of $b_1$ and $b_2$ in $LB$. If $O$ is the first (last) word in the sentence, we do not calculate any mean, and $bm$ will be simply equal to the frequency of $b_1$ ($b_2$). The process is repeated for the $M$ elements in $\mathscr{L}$. The values of $bm$ are combined with the cosine similarities for each $L_j$, to re-rank $\mathscr{L}$ as 
\begin{equation}
    \theta_j = \frac{\theta_j + bm_j}{2} \, ,  \,\,\,\,\, 1 \le j \le M \, .
\end{equation}
%
%

%

Finally, we take the word at the top of the list as the chosen candidate to substitute $O$. The idea is to select the word that is semantically most similar to $\vec{y}$, based on the analysis accomplished with \textit{Word2vec}, and maintain coherence of the text obtained with guidance of the language model. The process is repeated for each word in $f1$ and $f2$, except when we replace the last word in the second sentence, $w2_{L}$.
%
%
To replace $w2_{L}$, we present the word that substituted $w1_{L}$ in $f1$ as input to RIMAX. RIMAX returns a ranked list $LR$ with consonant and assonant rhymes related to $w1_{L}$. A score, $LR_w$ is attributed to each word, $w$ in $LR$, corresponding to a semantic similarity measure, which results from the semantic-phonetic analysis performed by a hybrid, automatic and  manual process. The scores are normalized in the interval $[0-1]$.
%
%
The scores in $LR$ are combined with the scores in $\mathscr{L}$. For this, an average score is calculated for each pair of elements $LR_w$ and $L_w$, where $L_w$ corresponds to the element of $\mathscr{L}$ with the information referring to $w$.
\begin{equation}
    \label{eq:combination}
    \theta_w=\frac{\theta_w+LR_w}{2}\, , \,\,\,\,\,\;\; \forall w \in LR \, .
\end{equation}


The words in $LR$ that do not exist in $\mathscr{L}$ are also considered, and since $LR_j$ is divided by 2, this strategy allows us to prioritize the elements contained in both lists. The new values in $\mathscr{L}$ are then processed with the language model as already described. Then we take the element in $\mathscr{L}$ in the first place, which contains the best semantic and coherent rhyme. 
%
%
Finally, a morphological analysis is performed once again with FreeLing, in order to convert the selected word into the correct inflection specified by the POS tag, if necessary. For that, we carry out conjugations and genre or number conversions.

\begin{figure}[htbp]
\centering
\includegraphics[width=9cm ]{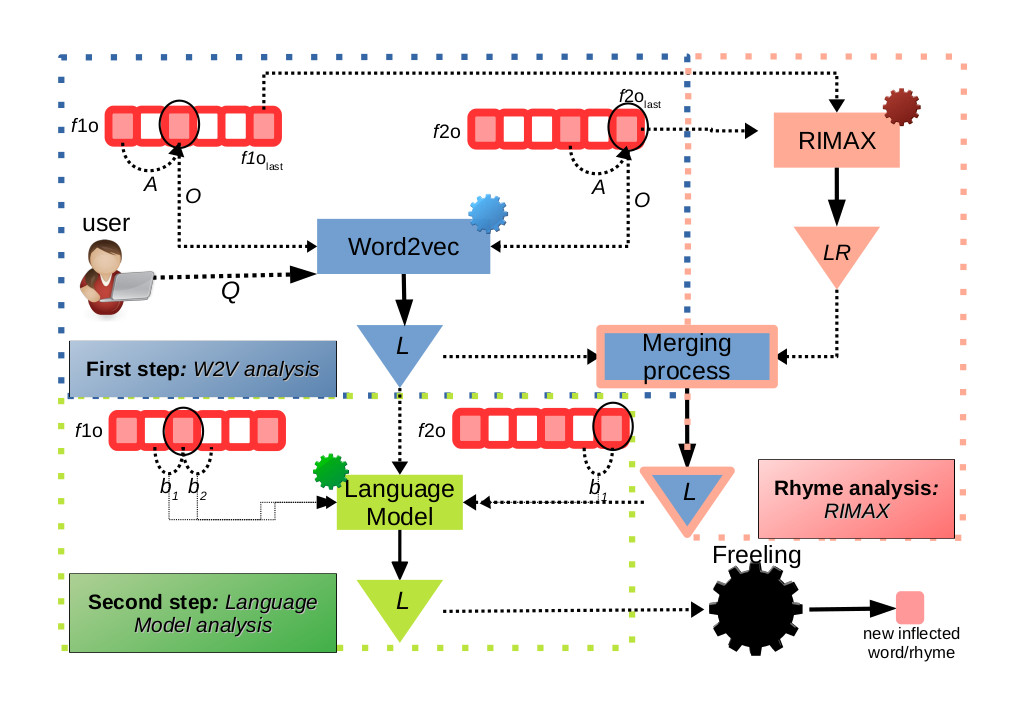}
\caption{Second Step: Vocabulary Selection}
\label{fig:vocSel}
\end{figure}

The result is a newly generated pair of phrases that does not exist in the \textbf{MegaLite-Es} corpus, where $w2_{L}$ must rhyme with $w1_{L}$. The model is illustrated in Fig. \ref{fig:vocSel}, where the two PGSs (for $f1$ and $f2$) can be appreciated at the top of the illustration. Both structures are sending inputs to the \textit{Word2vec} model, which receives $Q$, $A$ and $O$ in order to generate the list $\mathscr{L}$ with the vocabulary related to the inputs. It can be observed that the RIMAX module outputs its result to $f2$. RIMAX receives $w1_{L}$ and generates the list of rhymes $LR$, and then $\mathscr{L}$ and $LR$ are combined by Eq. (\ref{eq:combination}) to update the $\mathscr{L}$ list. The list $\mathscr{L}$ with the vocabulary is then sent to the language model module, for the selection of the most coherent option. Finally, the best option is processed with FreeLing, in order to make it respect the grammatical information provided by the POS tag in the PGS (to preserve inflection).
In the Fig. \ref{fig:vocSel}, we have marked in blue the \textbf{First step} where the semantic vocabulary list is generated to subsequently be sent to the \textbf{Second step}, the Language Model module marked in green. The pink section, the \textbf{Rhyme analysis}, is an independent section that is only executed once in the process.


We can illustrate an example in Spanish of our model as follows: 
\begin{enumerate}
    \item \textbf{Templates generation} (Canned Text):  \textit{Jamás he sido más [AQMS]; yo era ya un [NCMS].} / \textit{[VMI3S] el [NCMS] rápidamente hacia las [NCFP] que [VMII3P] el [NCMS].}
    \item \textbf{Vocabulary selection} (Word2vec):  \textit{Jamás he sido más [afectuoso]; yo era ya un [NCMS].} / \textit{[Corría] el [sol] rápidamente hacia las [cumbres] que [anteponían] el [NCMS].}
    \item \textbf{Rhymes selection} (RIMAX):  \textit{Jamás he sido más [afectuoso]; yo era ya un [\textbf{ofrecimiento}].} / \textit{[Corría] el [sol] rápidamente hacia las [cumbres] que [anteponían] el [\textbf{firmamento}].}
\end{enumerate}

\section{Experiments and Evaluation}
\label{sec:exp}

In preliminary tests of our proposal, we generated and evaluated 44 pairs of rhyming sentences that were generated using PGSs created from sentences in the \textbf{MegaLite-Es} corpus. The PGSs respect the rules specified in Section \ref{sec:FirstStep}. The new contexts are given by eleven different queries: \textit{amor, odio, tristeza, alegria, sol, luna, hombre, mujer, bosque, desierto, mar} (love, hate, sadness, joy, sun, moon, man, woman, forest, desert, sea). 
An examples of the generated sentences are listed below, where we show the queries, the sentences in Spanish in \textbf{bold} print, and in \textit{italic} their translation.

\begin{itemize}
    
    \item [\textit{sadness:}] \textbf{El sol de mediodía encapota sobre la inexpresable niebla de mi bosque.} | \textit{The midday sun overlays the inexpressible mist of my forest.}\\
    \textbf{Subía el sol rápidamente hacia las desolaciones que limitaban el zopilote.} | \textit{The sun was rising rapidly towards the desolations that bordered the vulture.}
    
\end{itemize}


Although general ATG tasks have been widely addressed by the research community, using different automatic evaluation protocols, it is difficult to implement automatic evaluation in the case of {\it literary text} due to the ambiguity and subjectivity involved in its interpretation and evaluation \cite{boden2004creative}. For this reason, we have performed a manual evaluation of our experiments, asking 6 people with a graduate degree in literature to evaluate the rhymes generated by our algorithm and their semantic relations. In previous general ATG models \cite{morenojimnez2020generacin,moreno_nldb}, criteria such as coherence and grammatical composition were evaluated. Here, we asked the evaluators to indicate if they perceived a rhyme between the last words of each sentence in a pair and also to specify their perception of the semantic relation between the two rhyming words, which could be one of the following: \textit{any relation}, \textit{low relation}, \textit{acceptable relation}, \textit{good relation} and \textit{strong relation}. 
We calculated the mode and median of the evaluator's feedback, obtaining a \textit{low relation} between the two rhyming words. This was to some extent expected because, when the model looks for the second rhyming word, the semantic analysis is performed considering not only the word to rhyme, but also the general context (the \textit{query}) and the adjacent word. For this reason, it is expected that, in some cases, the semantic relation between the two rhyming words cannot be preserved, although some relation was always perceived. For the evaluation of rhyme, we obtained encouraging results with a perception of rhymes in $61\%$ of the pairs of sentences.


\section{Conclusions and Future Work}
\label{sec:conclusion}

We have proposed a model capable of generating rhyming sentences in Spanish, although with a weak semantic relation between them. This can be improved by altering the semantic analysis, when selecting the second rhyme.
We have presented preliminary results showing that the model generates literary sentences that integrate semantic aspects with rhymes. Nevertheless, it must be considered that this task is still an open problem and further models, their extensions and generalizations, and experiments may confirm or improve the results that we have obtained.
We expect to perform more complex and extensive evaluations, and analyse more criteria, such as coherence, by generating more sentences and recruiting more evaluators. We also plan to conduct experiments in other languages, like French or Portuguese.


\noindent \textbf{Acknowledgment:}
We thanks CONACYT (Mexico), grant number 661101 and Agorantic (Avigon Université, France) by their financial support.

\bibliographystyle{plain}
\bibliography{sbc-template}

\end{document}